# Hybrid Digital Twin for process industry using Apros simulation environment

Mohammad Azangoo[1], Joonas Salmi[1], Iivo Yrjölä[1], Jonathan Bensky[1],
Gerardo Santillan[2], Nikolaos Papakonstantinou[3], Seppo Sierla[1], and Valeriy Vyatkin[1,4]

[1]*Department of Electrical Engineering and Automation, Aalto University, Espoo, Finland*
[2]*Semantum Oy, Espoo, Finland*
[3]*VTT Technical Research Centre of Finland Ltd, Espoo, Finland*
[4]*Department of Computer Science, Electrical and Space Engineering, Luleå University of Technology, Luleå, Sweden*
Email: {mohammad.azangoo, joonas.salmi, iivo.yrjola, jonathan.bensky}(at)aalto.fi,
gerardo.santillan(at)semantum.fi, nikolaos.papakonstantinou(at)vtt.fi, seppo.sierla(at)aalto.fi, valeriy.vyatkin(at)aalto.fi

*Abstract*—Making an updated and as-built model plays an important role in the life-cycle of a process plant. In particular, Digital Twin models must be precise to guarantee the efficiency and reliability of the systems. Data-driven models can simulate the latest behavior of the sub-systems by considering uncertainties and life-cycle related changes. This paper presents a step-by-step concept for hybrid Digital Twin models of process plants using an early implemented prototype as an example. It will detail the steps for updating the first-principles model and Digital Twin of a brownfield process system using data-driven models of the process equipment. The challenges for generation of an as-built hybrid Digital Twin will also be discussed. With the help of process history data to teach Machine Learning models, the implemented Digital Twin can be continually improved over time and this work in progress can be further optimized.

*Index Terms*—industry 4.0, automation, process industry, digital twin, machine learning, modeling, simulation, apros

## I. INTRODUCTION

There are different forms for system presentation and modeling, each form has individual features and advantages. A physics-based model is created based on known dynamics and understanding of the system's components. This requires detailed knowledge about how the components interact to create an accurate physics-based model. Data-driven models take a different approach, as they only look at inputs and outputs of a system. With enough scenarios and data collected, they are able to predict the system's behavior. A data-driven model can be a useful addition to a physics-based model in various situations. They often complement each other. A Digital Twin of a process plant is a digital representation of the physics and dynamics of the plant. Also, the Digital Twin is a live model which can communicate with the plant. Digital Twins use simulation models of physical systems which are developed using first-principles modeling or data-driven modeling. Hybrid Digital Twins use a combination of physics-based and data-driven models to create a more accurate tool. The proposed benefit of a hybrid model is that it might be simpler to create and might even be the only option if dynamics are unknown. It typically helps ensure that all aspects and variables are taken into account.

The data-driven models can be created by using Machine Learning, which is one of the most popular instruments in artificial intelligence. Machine Learning provides tools for capturing the system's behavior using its data history. It has progressed dramatically over recent decades, from laboratories to many commercial uses [1]. At the same time, Machine Learning has been playing an important role in process industries to identify new patterns of data and predict the behavior of systems more quickly and effectively. These supporting applications of Machine Learning have improved production methods in the process industry [2].

The objective of this paper is to develop a methodology for hybrid Digital Twin generation, as shown in Figure 1. Towards this objective, a Machine Learning model for a heater in the Aalto water process system was generated, and then a modeling interface to the Apros simulation environment was developed to link the build data-driven model to the physics-based Digital Twin. This paper presents the results of this work-in-progress on developing this methodology to generate hybrid Digital Twin for process systems using Machine Learning and Apros simulation software. The complete work report can be found in [3].

The rest of this paper is organized as follows, Section II will discuss related work, then Section III will present the main steps for hybrid Digital Twin generation. The step-by-step methodologies and also achieved results will be discussed in Section IV. The final remarks and future plans are provided in Section V.

## II. RELATED WORK

First principles and physics-based modeling are very important in the process industry. There are several applications for them in the process systems like monitoring, control system design, optimization and fault detection. Several attempts have been made by researchers to extract the physics-based Digital Twin from available sources of data in brownfield process systems like Piping and Instrumentation Diagrams (P&IDs) and 3D-scanned or computer-aided design (CAD) models of the systems [4]–[6].

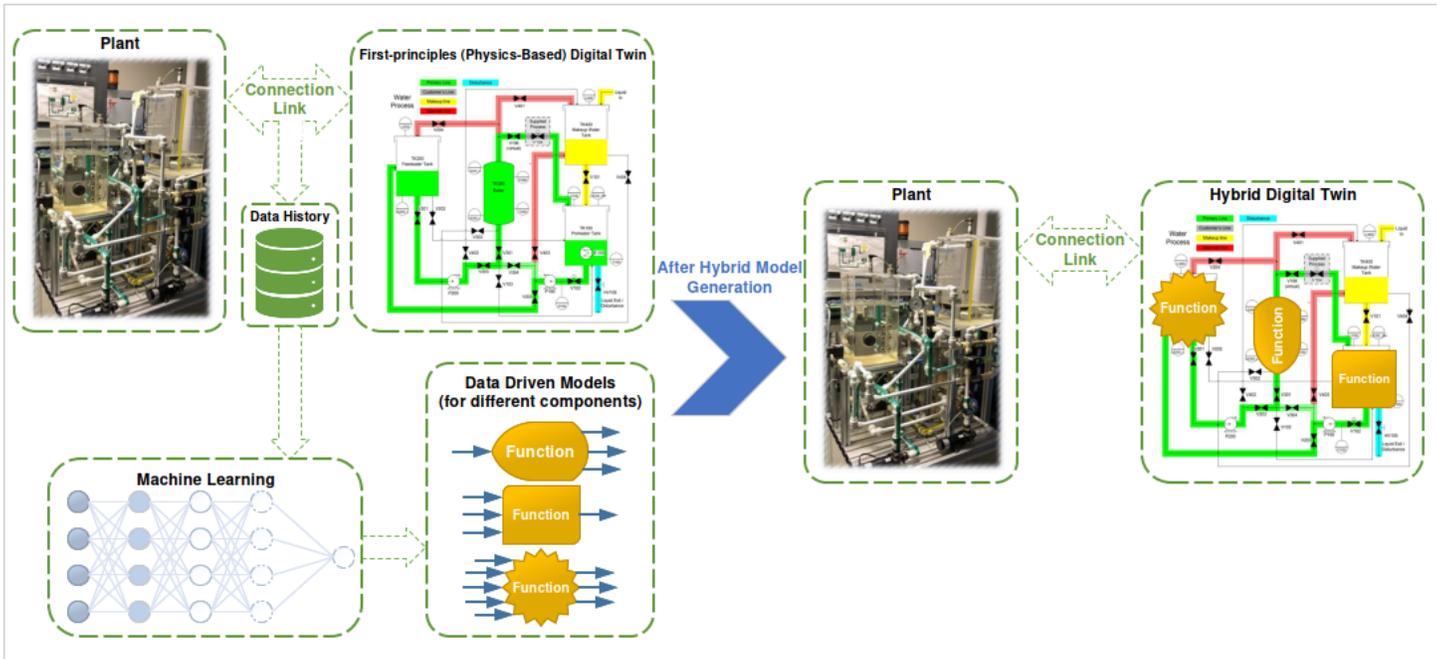

Fig. 1. The road map towards hybrid Digital Twin generation.

Machine Learning has been used widely in the process industry for different applications like data-driven model generation, data mining and analytics. There are four popular methods for using Machine Learning; while supervised learning and unsupervised learning are the most widely used Machine Learning methods in the industry, covering about 80–90 percent of all applications, the other two methods, semi-supervised learning and reinforcement learning, have scarcely been used in this domain [2]. Using Machine Learning related tools can improve the energy efficiency of the manufacturing and process industries, they can optimize systems by saving energy, reducing time and required resources, and also minimizing waste [7]. The concept of hybrid modeling has been used in the process industry for the last three decades. In the nineties, scientists first tried to empower first principles models based on prior knowledge by merging them with neural networks [8], [9]. Transition to Industry 4.0 concepts like big data and huge interests for digitization provided great infrastructure for the development of data-driven frameworks, and therefore data-related modeling approaches have become more popular these days [10]. However, to benefit more from the advantages of different types of modeling, it is better to combine them to create hybrid models. There are different structures for hybrid modeling like serial, parallel or surrogate methodologies; the best structure a hybrid model can be determined based on the requirements, the accuracy of current models and the amount of available data [10].

### III. PLANT DESCRIPTION AND HYBRID DIGITAL TWIN GENERATION

A simplified P&ID of the Aalto water process plant is shown in Figure 2. The plant consists of several tanks, pumps, a heater and a boiler which are connected by pipelines and equipped with sensors and actuators to control flow, temperature, pressure and surface levels. The plant features a heating element (E100) inside the pre-heater tank (Tank100) which can provide the hot water required for the virtual load (Valve106). For more information about Aalto water process plant you can see [11, pp. 12-23].

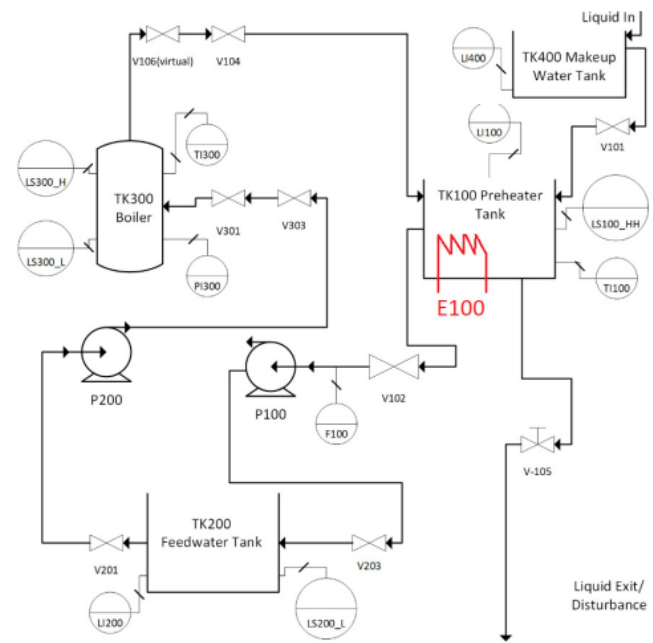

Fig. 2. A simplified P&ID of the Aalto water process plant, adapted from [11, p. 14].

Apros is a high-fidelity dynamic simulation software which can analyse dynamic behavior of process systems, automation and electrical systems. Apros software was used as the development environment, because of the available first principles model designed for Aalto water process system in the Apros environment [12]. Another benefit is that Apros has great support which makes the development and integration fairly easy. In addition, a major benefit is that the Apros tool is used globally in industries by many professionals. Apros has informative GUI and effective commands. It is fairly easy to learn and the tutorials provided by Apros makes it very easy to get into modeling and simulating. Apros has great materials which can be reached from Apros itself. This makes them very easy to access. Apros has support for Python libraries and Python language and we were able to create the data-driven model for Apros with Python and this was very helpful during integration time.

Data were collected from the Alto Water Process System to teach the Machine Learning model. The data from the real process was collected from the system PC using already set up communication between the system PLC and PC. In the system PC the data was available in Prosys OPC Historian Database which was connected to SQL database. From the SQL database we were able to export the data in CSV format and use it in Python. In Section I, the main steps of hybrid Digital Twin generation were presented briefly and the corresponding road map is shown in Figure 1, the progress on the implementation of the mentioned steps is discussed below:

- **Run physics-based Digital Twin to collect data:** Data history was collected from the physics-based Digital Twin and the real system.
- **Make Machine Learning Models for process components:** The Machine Learning model was generated for the heater based on collected data.
- **Interface for Data-driven models replacement:** The interface for model modification was implemented in Apros.
- **Hybrid model generation:** The new Machine Learning model for the heater was replaced in the physics-based Digital Twin.
- **Evaluation and Digital Twin commissioning:** The generated hybrid Digital Twin was tested and evaluated under operating conditions.

## IV. INITIAL RESULTS

The purpose of the work was to increase the use of data-driven modeling in the building of Digital Twins, as traditionally physics-based modeling is more common. To achieve this, we developed a data-driven model for the heating component of the Aalto water process system. In this way we replaced one of the components of the overall digital model with a data-driven model. To manage this, we had to first collect data from the physical plant, then create an integration between Apros and Python, which we used to create a Machine Learning model. Once these steps were completed, we could implement our training model and use this data-driven component as a part of our new live simulation model of the water process plant.

In order to speed up the process, we used the physics-based Apros model to collect data for the data-driven model. This was done in order to have enough time to test the Python-Apros integration and to be able to evaluate the whole project. It will be relatively easy to retrain our Machine Learning model in the future to improve performance. The following subsections will discuss the main challenges in more detail.

### A. Integration, Setup and Results

For the model integration, we set up the connection in Apros. It has comprehensive equipment libraries and support for process simulation. Also, Apros version 6.10.32 can support a Python interface for modeling modifications, which was used in this project. For the Apros-Python integration, first we needed to set up the Python integration library by installing Python semantics in Apros. Then we had to create a new *User Component*, where we could implement the Python script. The *User Component* includes the following components: Configuration, symbol, Terminals, *Initalization_script*, *MLModelHeat_script* and *python_module*. In the terminals component all the terminals should be changed according to the features which are selected to the model. Figure 3 shows the implemented *User Component* and its required connections, which can be used for the replacement of data-driven models.

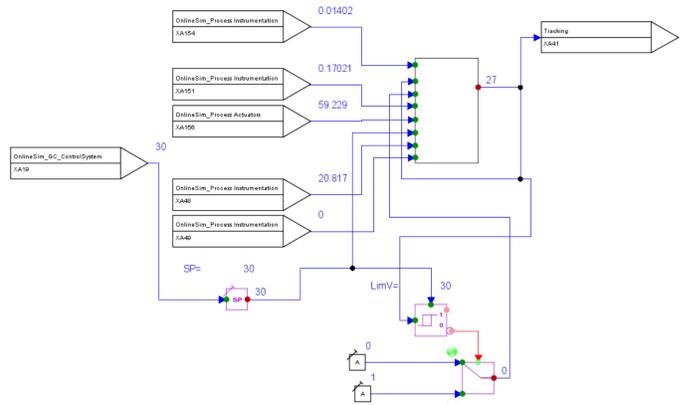

Fig. 3. Integration-Data-driven model diagram

### B. Machine Learning Model

To develop our Machine Learning model we used an open source Python library called Keras [13], which is built on top of TensorFlow [14], an end-to-end open platform. At the design phase, we found recent studies where deep learning was used to predict dynamical nonlinear systems' behavior [15], [16]. Based on these studies, and other literature, we decided to use Long Short-Term Memory network (LSTM)-based sequence-to-sequence recurrent neural networks [17] to generate the data-driven model. Recurrent neural networks have been used for presenting data sequences like time series data but because it uses back propagation through time it has a vanishing gradient problem. LSTM avoids this problem using gated units [15].

The Machine Learning model consists of an encoder part that was designed to observe the internal state of the system based on previous feature values, and a decoder part that predicts the label value based on the state of the encoder and current feature values. Next, the features for the Machine Learning model were chosen based on basic knowledge of physics and intuition about the main parameters of the system that affects the temperature of tank T100. The final Machine Learning model was plotted using *keras.utils.plot_model()* function and can be seen in Figure 4. As an optimizer we used ready-made *RMSprop* from Keras library and the Mean Squared Error loss function.

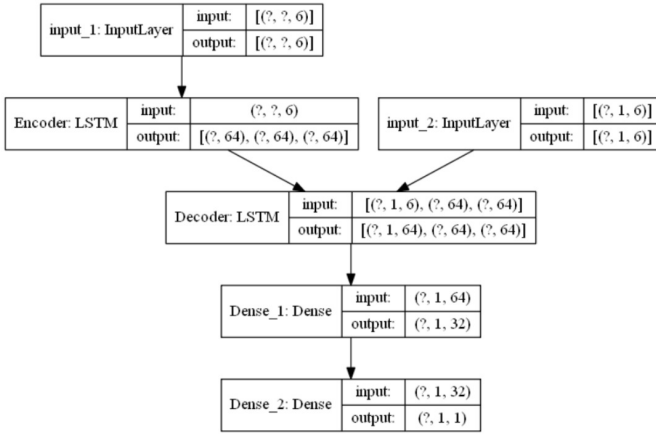

Fig. 4. LSTM, Implemented Machine Learning model.

## C. Hybrid Digital Twin Behavior

In Figure 5 the dynamic performance of the data-driven model is compared to the physics-based model for temperature of Tank100 in a set point change scenario.

However the generated Machine Learning model has quite a lot of oscillation, but the average value tracks the system's value very well and it can keep up when the set point is changed. The data-driven model has the same rise speed as the system.

Next we will present the results from the tracking simulation where we adjusted the simulation state according to the Machine Learning model, as shown in Figure 6. We could see the slight changes in the process instruments and the accuracy improvement when the Machine Learning model is in the loop. The overall result shows that the Machine Learning model is able to predict and follow the process changes.

## D. Discussion

This work proved that it is possible to create a functional data-driven model in Apros and integrate it with first-principle models to get a hybrid model and a hybrid Digital Twin of the system. As this process is very well understood with physics-based models, it was unlikely that our Machine Learning model would achieve the accuracy of the original Apros model. Also, this project shows the possibility of adjusting the simulation according to the data-driven model.

We used an LSTM decoder/encoder method to generate the Machine Learning model. As with other Machine Learning models, this version also needed a large amount of data to give more accurate results. However, it was time-consuming to collect data and set up the Machine Learning model. In order to speed up the checks of the Machine Learning model and verify that it worked as expected, we used Apros simulation data for training purposes. The results show a Machine Learning model can predict non-linear dynamic behavior in some sense. This should be improved by collecting more data for teaching. A Machine Learning model is a "plug in" module in the presented solution, and it can be improved going forward.

Another difficulty that we encountered was the simulation sampling rate. The selected Machine Learning model uses a fixed time-step which means that we would like to have the same steps in the simulation. The model uses one second time-steps and we tried to use that time-step in our simulation, but Apros was not able to keep a fixed sampling rate through the simulation. The rate fluctuated between 0.5 and 1.0 seconds in Apros. We assume that this affected the prediction as well as the simulation.

## V. CONCLUSION & FUTURE WORKS

As was shown in this paper, data-driven models can be used to understand phenomena and accurately predict and simulate outcomes. The introduced hybrid Digital Twin concept will be useful, especially in situations where physical modeling is too complex or time-consuming, but we are able to measure inputs and outputs successfully. The concrete contribution of this paper was to show how to use the first-principles model of a process plant and replace one unit of that model with a data-driven model. The heating component of the Aalto water process plant was replaced with a data-driven model. This was done by creating a Machine Learning model on Python, integrating it with the simulation modeling software Apros.

There are multiple possibilities for improving the presented solution. It would simplify the procedure if data could be collected from the real system automatically. Also, the more advanced Machine Learning model should be developed further and the features could be adjusted. The next step after these improvements would be to replace more parameters or even a whole component with data-driven models. To be able to successfully implement a whole component, there will be more problems to be solved in data collection, feature selection and implementation. Also, a real-time upgrade of a Digital Twin could be achieved by making a live connection between database, Machine Learning core and simulation software.

## REFERENCES

[1] M. I. Jordan and T. M. Mitchell, "Machine learning: Trends, perspectives, and prospects," *Science*, vol. 349, no. 6245, pp. 255–260, 2015. [Online]. Available: https://science.sciencemag.org/content/349/6245/255
[2] Z. Ge, Z. Song, S. X. Ding, and B. Huang, "Data mining and analytics in the process industry: The role of machine learning," *IEEE Access*, vol. 5, pp. 20 590–20 616, 2017.

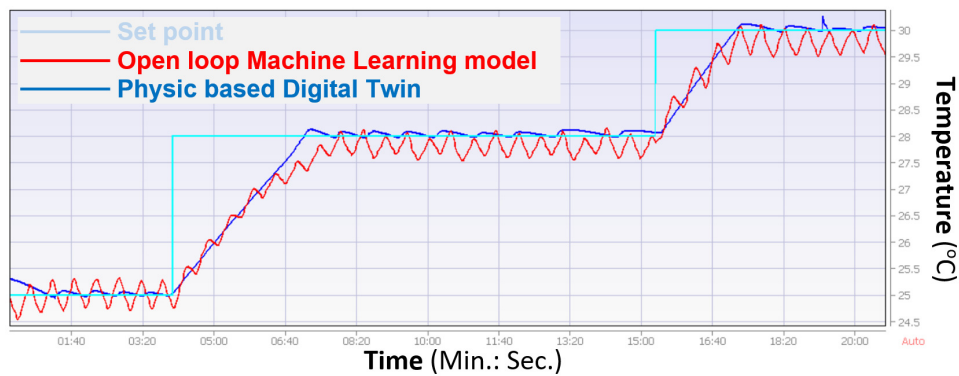

Fig. 5. Open-loop behavior of data-driven model vs. physics-based Digital Twin

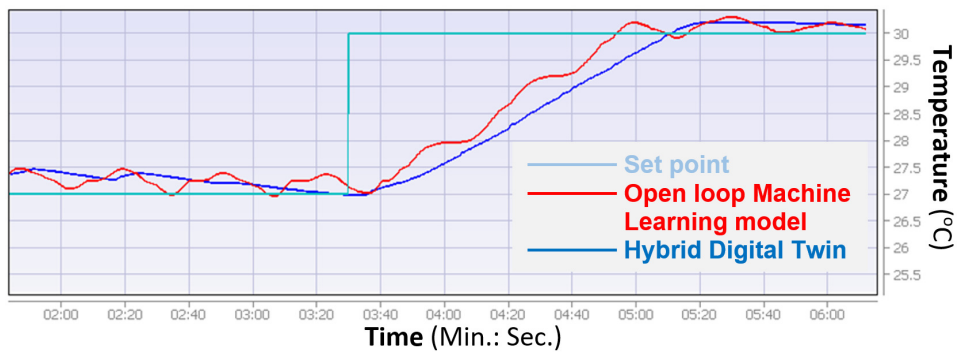

Fig. 6. Hybrid Digital Twin and open-loop Machine Learning model behavior

[3] J. Salmi, J. Bensky, and I. Yrjölä, "Hybrid digital twin for process industry." [Online]. Available: https://wiki.aalto.fi/display/AEEproject/Hybriddigitaltwinforprocessindustry
[4] S. Sierla, M. Azangoo, A. Fay, V. Vyatkin, and N. Papakonstantinou, "Integrating 2d and 3d digital plant information towards automatic generation of digital twins," in *2020 IEEE 29th International Symposium on Industrial Electronics (ISIE)*, 2020, pp. 460–467.
[5] S. Sierla, M. Azangoo, and V. Vyatkin, "Generating an industrial process graph from 3d pipe routing information," in *2020 25th IEEE International Conference on Emerging Technologies and Factory Automation (ETFA)*, vol. 1, 2020, pp. 85–92.
[6] S. Sierla, L. Sorsamäki, M. Azangoo, A. Villberg, E. Hytönen, and V. Vyatkin, "Towards semi-automatic generation of a steady state digital twin of a brownfield process plant," *Applied Sciences*, vol. 10, no. 19, 2020. [Online]. Available: https://www.mdpi.com/2076-3417/10/19/6959
[7] D. A. Narciso and F. Martins, "Application of machine learning tools for energy efficiency in industry: A review," *Energy Reports*, vol. 6, pp. 1181–1199, 2020. [Online]. Available: https://www.sciencedirect.com/science/article/pii/S2352484719308686
[8] D. C. Psichogios and L. H. Ungar, "A hybrid neural network-first principles approach to process modeling," *AIChE Journal*, vol. 38, no. 10, pp. 1499–1511, 1992, cited By :486.
[9] H.-T. Su, N. Bhat, P. Minderman, and T. McAvoy, "Integrating neural networks with first principles models for dynamic modeling," in *Dynamics and Control of Chemical Reactors, Distillation Columns and Batch Processes*, ser. IFAC Symposia Series, J. BALCHEN, Ed. Oxford: Pergamon, 1993, pp. 327 332. [Online]. Available: https://www.sciencedirect.com/science/article/pii/B9780080417110500544
[10] J. Sansana, M. N. Joswiak, I. Castillo, Z. Wang, R. Rendall, L. H. Chiang, and M. S. Reis, "Recent trends on hybrid modeling for industry 4.0," *Computers & Chemical Engineering*, vol. 151, p. 107365, 2021. [Online]. Available: https://www.sciencedirect.com/science/article/pii/S0098135421001435
[11] T. Lopez, "Modeling a Laboratory Scale Heating Process in a Machine Readable Format," Master's thesis, Aalto University. School of Electrical Engineering, 2019. [Online]. Available: http://urn.fi/URN:NBN:fi:aalto-201910275841
[12] G. S. Martínez, S. Sierla, T. Karhela, and V. Vyatkin, "Automatic generation of a simulation-based digital twin of an industrial process plant," in *IECON 2018 - 44th Annual Conference of the IEEE Industrial Electronics Society*, 2018, pp. 3084–3089.
[13] F. Chollet *et al.*, "Keras," https://keras.io, 2015.
[14] M. Abadi, P. Barham, J. Chen, Z. Chen, A. Davis, J. Dean, M. Devin, S. Ghemawat, G. Irving, M. Isard, M. Kudlur, J. Levenberg, R. Monga, S. Moore, D. G. Murray, B. Steiner, P. Tucker, V. Vasudevan, P. Warden, M. Wicke, Y. Yu, and X. Zheng, "Tensorflow: A system for large-scale machine learning," in *Proceedings of the 12th USENIX Conference on Operating Systems Design and Implementation*, ser. OSDI'16. USA: USENIX Association, 2016, p. 265–283.
[15] J. Gonzalez and W. Yu, "Non-linear system modeling using lstm neural networks," *IFAC-PapersOnLine*, vol. 51, no. 13, pp. 485 489, 2018, 2nd IFAC Conference on Modelling, Identification and Control of Nonlinear Systems MICNON 2018. [Online]. Available: https://www.sciencedirect.com/science/article/pii/S2405896318310814
[16] A. Wagh, "Deep learning of nonlinear dynamical system," *Open Access Master's Thesis, Michigan Technological University*, 2020.
[17] S. Hochreiter and J. Schmidhuber, "Long short-term memory," *Neural Comput.*, vol. 9, no. 8, p. 1735–1780, Nov. 1997. [Online]. Available: https://doi.org/10.1162/neco.1997.9.8.1735